\documentclass[conference]{IEEEtran}
\IEEEoverridecommandlockouts
\usepackage{cite}
\usepackage{amsmath,amssymb,amsfonts}
\usepackage{algorithmic}
\usepackage{graphicx}
\usepackage{textcomp}
\usepackage{xcolor}
\usepackage{dirtytalk}
\usepackage{multirow}
\usepackage{url}
\def\BibTeX{{\rm B\kern-.05em{\sc i\kern-.025em b}\kern-.08em
    T\kern-.1667em\lower.7ex\hbox{E}\kern-.125emX}}
\begin{document}

\title{Enhanced Labeling Technique for Reddit Text and Fine-Tuned Longformer Models for Classifying Depression Severity in English and Luganda\\

\thanks{This research was supported by the Institute for Information \& communications Technology Promotion(IITP) grant funded by the Korea government(MSIT) (No. 2018-0-00749, Development of virtual network management technology based on artificial intelligence), Korea International Cooperation Agency (KOICA), and  by Basic Science Research Program through the National Research Foundation of Korea(NRF) funded by the Ministry of Education(NRF-2022R1A2C1012633).}
}

\author{
\IEEEauthorblockN{Richard Kimera}
\IEEEauthorblockA{\textit{Department of Advanced Convergence} \\
\textit{Handong Global University}\\
Pohang, South Korea \\
kimrichies@handong.ac.kr}
\and
\IEEEauthorblockN{Daniela N. Rim}
\IEEEauthorblockA{\textit{School of Computer Science and
Electrical Engineering} \\
\textit{Handong Global University}\\
Pohang, South Korea \\
danielarim@handong.ac.kr}
\and
\IEEEauthorblockN{Joseph Kirabira}
\IEEEauthorblockA{\textit{Department of Psychiatry} \\
\textit{Busitema University (Faculty of Health Sciences)}\\
Mbale, Uganda \\
jkirabira.fhs@busitema.ac.ug}
\and
\IEEEauthorblockN{Ubong Godwin Udomah}
\IEEEauthorblockA{\textit{Department of Psychiatry} \\
\textit{University of Uyo teaching hospital}\\
Uyo, Akwa Ibom, Nigeria \\
bloominasant@yahoo.com}
\and
\IEEEauthorblockN{Heeyoul Choi}
\IEEEauthorblockA{\textit{School of Computer Science and
Electrical Engineering} \\
\textit{Handong Global University}\\
Pohang, South Korea \\
hchoi@handong.edu}
}

\maketitle

\begin{abstract}
Depression is a global burden and one of the most challenging mental health conditions to control. Using the BDI questionnaire, experts can detect its severity early, administer appropriate medication to patients, and impede its progression. Owing to the fear of potential stigmatization, many patients turn to social media platforms such as Reddit for advice and assistance at various stages of their journey. This research extracts text from Reddit to facilitate the diagnostic process, employs a proposed labelling approach to categorize the text, and subsequently fine-tunes the Longformer model. The model’s performance is compared against the baseline models, including Naive Bayes, Random Forest, Support Vector Machines, and Gradient Boosting. Our findings reveal that the Longformer model outperforms the baseline models in both English (48\%) and Luganda (45\%) languages on a custom-made dataset.\\

\end{abstract}

\begin{IEEEkeywords}
Depression Severity, Longformer, BART, fine-tuning, Luganda, Reddit 
\end{IEEEkeywords}

\section{Introduction}
Mental health disorders are prevalent worldwide and are predicted to be the leading cause of disease burden by 2030 \cite{fogarty2021physical}. Amongst them, \textit{major depressive episodes} (depression) is a common psychiatric condition that can be challenging to manage due to its various presentations, unpredictable course and prognosis, and variable response to treatment \cite{mcallister2020identification}. Access to professional mental health assessment, care, and resources is often limited to a general  demographic. First, detecting a patient's severity level is important to provide proper medication and prevent its advancement to stages that can lead to suicide tendencies and death. To this end, when diagnosing depression, doctors use the Beck Depression Inventory (BDI) questionnaire. This 21-question multiple-choice has a set of four (4) possible choices for each question, ranging in severity. It can be self-administered, and the obtained score will determine a depression severity categorized into either of the six; \textit{normal}, \textit{mild}, \textit{moderate}, \textit{borderline}, \textit{severe}, and \textit{extreme} \cite{beck1987beck}. This resource-intensive task is hard to administer, especially in resource-constrained settings and societies where mental health patients are stigmatized. With this fear, many patients resort to using social media to anonymously share and receive global feedback from people with similar conditions, recovering patients, and experts. One example is Reddit\footnote{https://www.reddit.com/}, a platform allowing users to post and exchange ideas freely. This social media platform has over 57 million daily active users, with over 100,000 active communities. It also has over 50,000 daily active moderators \cite{redditinc}. The modulation can be done by moderators, administrators and a modulation tool (AutoMod), and if the content violates the platform terms of use, it is removed. In 2022 3.7\% of the total content created was removed \cite{reddittransparency}.

Text collected from Reddit for types of depression classification purposes is usually labelled by experts, trained personnel \cite{kim2022analysis}, or according to the corresponding category of the mental illness it is associated with \cite{kim2020deep, murarka2021classification}. This classified text is then used to train machine learning or deep learning models, an approach that is not new. \cite{naseem2022early}. Razavi et al. \cite{razavi2020depression} used the BDI-II to measure the severity of depression on a wide array of machine learning classification algorithms. Similarly, BDI-II was used in a study exploring the diagnostic ability of three machine learning methods for evaluating the depression status of Chinese recruits \cite{francese2023emotion}. 

Transformer based models such as BERT \cite{vaswani2017attention,devlin2018bert} have been fine-tuned on Reddit text for classification tasks \cite{shounak2022reddit, caselli2020hatebert}. \cite{gabin2021multiple} modified BERT for Multiple Choice Question Answering (MCQA) to predict users' answers to the BDI-II questionnaire. Variants such as RoBERTa \cite{liu2019roberta} have been used to classify mental health disorders such as depression, anxiety, bipolar disorder, Attention Deficit Hyperactivity Disorder, and Post Traumatic Stress Disorder\cite{murarka2021classification}. Using pre-trained models can maximize data efficiency, allowing for effective fine-tuning on smaller task-specific datasets. One of the setbacks of models that use the original attention mechanism is that they are limited to handling a maximum of 512 tokens. This is because the self-attention mechanism scales quadratically. The Longformer model was invented to solve this limitation. It uses the local window attention to scale linearly and global attention to attend to the entire sequence \cite{beltagy2020longformer}. It can handle eight times longer tokens than BERT, an attribute necessary to process longer texts (e.g., from Reddit) to detect severity levels more precisely. The Longformer model has been used in clinical text and outperforms clinical-BERT and clinical-Big Bird models \cite{li2023comparative}. It has been used to detect depression in users from web-based forums \cite{owen2023enabling}, and predicts differential responses to antidepressant classes using electronic health records \cite{sheu2023ai}.

Most of these models have been trained in English, leaving low-resource languages like Luganda unattended. Luganda is one of the morphologically rich Bantu languages spoken in Uganda by over half of the population and neighboring East African countries. Funding for mental health services in Uganda remains low by international standards, with only 1\% of GDP allocated for mental health services \cite{iversen2021child}. This lack of resources has limited access to mental health services, particularly in rural areas\cite{newman2022access}. Furthermore, language and cultural norms influence communication about mental health topics experienced by patients receiving mental health treatment\cite{dagsvold2015can}. Therefore, in such a multilingual society, it is important to provide various avenues to handle mental health problems, particularly depression.
\\ \\
In this research, our objectives encompass two main aspects; 
\begin{enumerate}
    \item introducing a method for labelling social media text through a combination of keyword matching, a context-aware BART model, and an expert, and 
    \item refining the Longformer model's performance in classifying depression severity using fine-tuning, while focusing on both English and Luganda languages.
\end{enumerate}

\section{Methods}

\subsection{Data collection}\label{AA}
Using the PRAW API, we collected  a total of 1807 sentences from the r/depression subreddit\footnote{https://www.reddit.com/r/depression/}. The community ranks 805 with 972,203 subscribers \cite{subredditstats}. It was assumed that the majority of the people who post under this subreddit have had a long-standing period of depression, as seen in the sample text below [sic];

\say{\textit {I feel like a complete failure. I can’t hold down a job and am getting a medical withdrawal from my semester in university. I feel like a waste of time and so much wasted effort for this school semester plus all the money. I’m having the hardest time doing basic self care activities like showering and changing my clothes. I feel so incredibly isolated and yet I keep ignoring peoples texts. I do not know what to do and I am sleeping around 12 hrs a night. I feel very very hopeless. Was just diagnosed with major depressive disorder and c-ptsd. Don’t know where to go from here but trying to hold on.}}\\


The collected text was pre-processed to input for the models. First, every paragraph was converted to lowercase words. Then, stop words, unnecessary punctuations, spaces, and hyperlinks were removed.   

\subsection{Data labelling}\label{AA}
\begin{enumerate}

\item \textbf{Keyword extraction and matching.}\\
  Using NLTK\cite{loper2002nltk}, a list of keywords were extracted from the 21 questions of the BDI questionnaire, while retaining the corresponding scores ranging from 0 to 3. A score of 0 indicates the absence of a symptom, while a score of 3 indicates the most severe manifestation of a symptom.
  
  Pattern matching was then conducted on each of the extracted sentences. A score value was assigned for every word match and aggregated to calculate a total score. The label was assigned based on the score range, as depicted in Figure \ref{fig:keywords}, following the original BDI scoring standard.\\
  \begin{figure}[ht]
    \centering
    \begin{minipage}{0.48\textwidth}
        \centering
        \includegraphics[width=\linewidth]{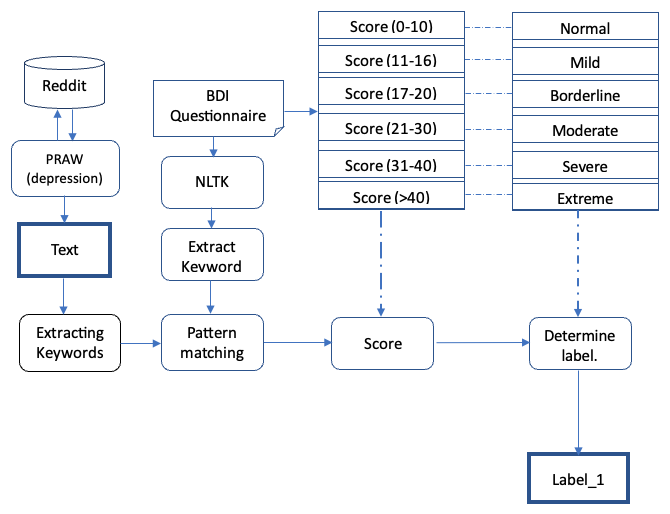}
        \caption{Keyword extraction and matching.}
        \label{fig:keywords}
    \end{minipage}
  \end{figure}

\item \textbf{Classification with BART.}
  In this approach (Figure \ref{fig:bard}), we supplied the extracted text and the BDI severity labels to a pre-trained transformer-based BART model \cite{lewis2019bart}. This model includes scientific text and medical content in its training stage.\\
  
  \begin{figure}[ht]
    \centering
    \includegraphics[width=0.48\textwidth]{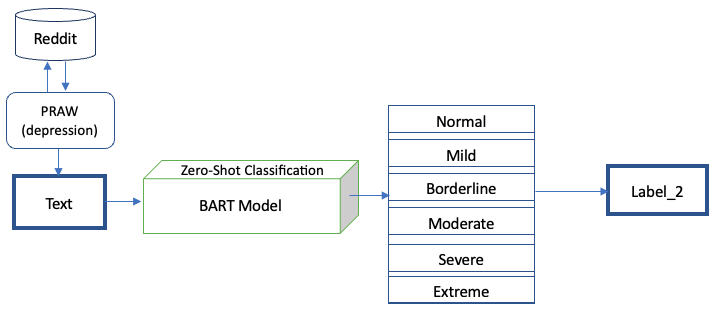}
    \caption{BART model for labelling.}
    \label{fig:bard}
    \end{figure}
  
  \item \textbf{Domain expert.}
  
  This labelling was performed by a Psychiatrist who assigned the labels considering the BDI questionnaire and their expertise (Figure \ref{fig:expert}).\\ 
  
  \begin{figure}[ht]
    \centering
    \includegraphics[width=0.48\textwidth]{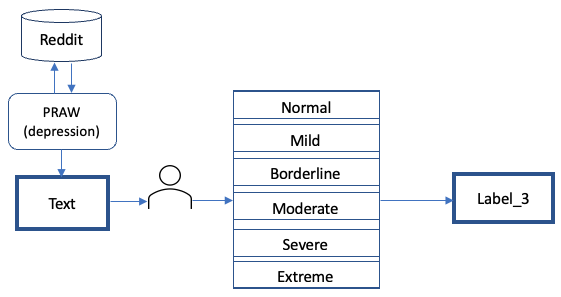}
    \caption{Expert manual labelling.}
    \label{fig:expert}
    \end{figure}

\item \textbf{Selecting the final label.}
    
    The final label for the input text was assigned by weighted majority voting. The labels from \textbf{1)}, \textbf{2)} and \textbf{3)} were aggregated; the label was immediately assigned if the three agreed. If only two agreed, then that was the resulting label, and if none of them agreed, the final label was the expert's (Figure \ref{fig:fin}). 
    
    \begin{figure}[ht]
    \centering
    \includegraphics[width=0.48\textwidth]{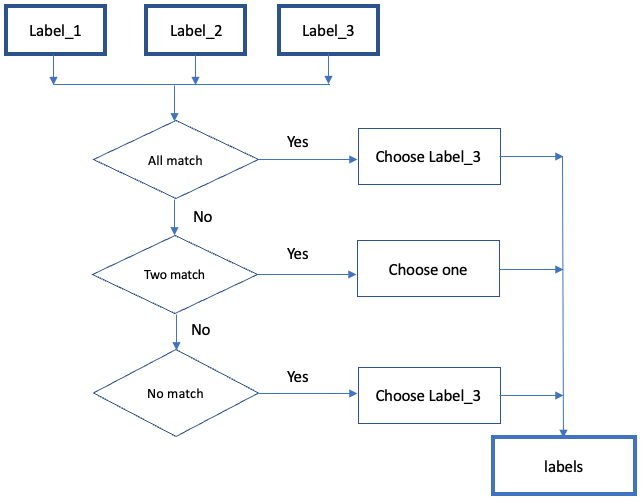}
    \caption{Final label selection using weighted majority voting. }
    \label{fig:fin}
    \end{figure}
\end{enumerate}

Finally, we obtained a labelled dataset with six classes: normal, mild, borderline, moderate, severe, and extreme. However, the "borderline" class was merged with the "mild" class labels, and the "extreme" class was merged with the "severe" class due to a low sample count of 17 and 23, respectively. Table \ref{tab: My table} shows the final class distribution. 

\begin{table}[htbp]
\caption{Data distribution for the final synthetic labelled dataset, for Validation (V) and Testing (T).}
\begin{center}
\begin{tabular}{cccc}
\hline
\textbf{Label}& 
{\textbf{Total}} &
{\textbf{English(V/T)}}&
{\textbf{Luganda(V/T)}}
\\
\hline
Normal & 301 & 12/14 & 12/11\\
Mild & 255  & 17/16 & 12/12\\
Moderate & 372 & 15/14 & 21/21 \\
Severe & 215  & 13/14 & 12/14 \\
\hline
\end{tabular}
\label{tab: My table}
\end{center}
\end{table}

\subsection{Classification models}\label{AA}
We used Google\footnote{https://translate.google.com/} to translate the input text from English to Luganda. The dataset was shuffled for each language and divided into training, testing, and validation. The training set was oversampled with the synthetic minority over-sampling technique (SMOTE) algorithm to address class imbalances \cite{chawla2002smote}.

For the final classification, we fine-tuned the Longformer model. This model has the ability to handle longer sequences of more than 512 tokens. It adjusts its attention mechanism for long sequences by combining strategies to lessen the computational constraints of processing such sequences. It attends to key parts of the input sequence using a combination of global and local attention methods, allowing it to capture long-range dependencies of the input sequence. It also employs an inter-attention mechanism that computes a distinct representation of the input for each output step, allowing the decoder to effectively ``look at" the input's relevant part(s) for each output step. As a result, the encoder is relieved of the burden of encoding all information about the input sequence into a fixed-size rich representation vector. It has performed well in various tasks such as machine reading comprehension, summarization, and question answering \cite{basafa2021nlp}.

We used the Longformer\footnote{https://huggingface.co/allenai/longformer-base-4096} model provided by the HuggingFace library \cite{jain2022hugging}. The model consists of a default configuration of a hidden size of 768, 12 attention heads, an attention window size of 512, and 12 layers. The average number of characters per input varied between both languages, with English comprising 893 characters and Luganda 1776.

To optimize the model, we performed a hyperparameter search involving 8, 16, and 32 batch sizes, and learning rates ranging from 1e-1 to 5e-1. Additionally, early stopping was implemented. Model performance evaluation on validation and testing sets was based on precision, recall, F1-score and accuracy metrics.

After converting our data using TF-IDF, the following machine learning models were employed as baseline models: Naïve Bayes (NB) classifier, Random Forest (RF), Support Vector Machines (SVM), and Gradient Boosting (GB).

\section{Results}\label{AA}
The models were trained on Luganda and English,  separately, and validated. Subsequently, they were evaluated using the test dataset (Table \ref{tab: My table}). Precision, recall, and F-1 score were calculated for each severity level. The overall accuracy was calculated for each of the models employed (Table \ref{tab:Engres} and Table \ref{tab:Lugres}).

Regarding the English dataset in Table \ref{tab:Engres}, the Random Forest model achieved the highest accuracy among the baseline models on the testing set (43\%). The gradient boosting model  failed as it could not predict any of the samples in the severe class. SVM had the lowest performance (41\%).
On the other hand, the Longformer model outperformed the baseline models with an accuracy of 48\%. The hyperparameters used included a batch size of 16, a learning rate of 5e-5, and a dropout rate of 0.1.

For the Luganda dataset, similar to English, the random forest achieved the highest accuracy (40\%), while SVM and GB performed poorly with failure to predict any of the samples in Mild and Severe classes respectively. However, unlike English, Naive Bayes achieved the lowest value (34\%).
The fine-tuned Longformer model achieved the highest accuracy of 45\%, as shown in Table \ref{tab:Lugres}. The hyperparameters used were a batch size of 16, and a learning rate of 4e-4.
The performance did not require a dropout layer, as its inclusion negatively impacted the experiments.

\begin{table}
\centering
\caption{Performance Metrics for Different Models on English}
\begin{tabular}{lcccccc}
\hline
& \multicolumn{5}{c}{Model} \\
\hline
Class & Metric & NB & RF & SVM & GB & Longformer\\
\hline

\multirow{5}{*}{Normal} 
& Precision & 1.00 & 0.69 & 0.10 & 0.58 & 1.00 \\
& Recall    & 0.50 & 0.79 & 0.50 & 0.50 & 0.57\\
& F-1       & 0.67 & 0.73 & 0.67 & 0.54 & 0.73\\
\hline

\multirow{5}{*}{Mild} 
& Precision & 0.23 & 0.45 & 0.50 & 0.28 & 0.37\\
& Recall    & 0.19 & 0.31 & 0.12 & 0.44 & 0.69\\
& F-1       & 0.21 & 0.37 & 0.20 & 0.34 & 0.48 \\
\hline

\multirow{5}{*}{Moderate} 
& Precision & 0.15 & 0.27 & 0.28 & 0.24 & 0.34\\
& Recall    & 0.14 & 0.57 & 0.79 & 0.36 & 0.43\\
& F-1       & 0.15 & 0.36 & 0.21 & 0.29 & 0.39\\
\hline

\multirow{5}{*}{Severe} 
& Precision & 0.44 & 0.10 & 0.57 & 0.00 & 1.00\\
& Recall    & 0.79 & 0.07 & 0.29 & 0.00 & 0.21\\
& F-1       & 0.56 & 0.13 & 0.38 & 0.00 & 0.35\\

\hline
\multirow{1}{*}{Accuracy} 
&           & 0.40 & 0.43 & 0.41 & 0.33 & 0.48\\
\hline
\multicolumn{4}{l}
{Performance Analysis by Severity Level}\\
\label{tab:Engres}
\end{tabular}
\end{table}

\begin{table}
\centering
\caption{Performance Metrics for Different Models on Luganda}
\begin{tabular}{lcccccc}
\hline
& \multicolumn{5}{c}{Model} \\
\hline
Class & Metric & NB & RF & SVM & GB & Longformer\\
\hline

\multirow{5}{*}{Normal} 
& Precision & 1.00 & 0.43 & 1.00 & 0.22 & 1.00 \\
& Recall    & 0.45 & 0.83 & 0.67 & 0.92 & 0.55\\
& F-1       & 0.62 & 0.57 & 0.80 & 0.35 & 0.71\\
\hline

\multirow{5}{*}{Mild} 
& Precision & 0.07 & 0.27 & 0.00 & 0.50 & 0.30\\
& Recall    & 0.08 & 0.25 & 0.00 & 0.17 & 0.50\\
& F-1       & 0.08 & 0.26 & 0.00 & 0.25 & 0.37 \\
\hline

\multirow{5}{*}{Moderate} 
& Precision & 0.38 & 0.41 & 0.47 & 0.33 & 0.45\\
& Recall    & 0.29 & 0.43 & 0.90 & 0.05 & 0.43\\
& F-1       & 0.32 & 0.42 & 0.62 & 0.08 & 0.44\\
\hline

\multirow{5}{*}{Severe} 
& Precision & 0.35 & 1.00 & 0.86 & 0.00 & 0.42\\
& Recall    & 0.57 & 0.08 & 0.50 & 0.00 & 0.36\\
& F-1       & 0.43 & 0.15 & 0.63 & 0.00 & 0.38\\

\hline
\multirow{1}{*}{Accuracy} 
&           & 0.34 & 0.38 & 0.40 & 0.24 & 0.45 \\
\hline
\multicolumn{4}{l}
{Performance Analysis by Severity Level}\\
\label{tab:Lugres}
\end{tabular}
\end{table}

\section{conclusion and future work}\label{AA}
With our success in fine-tuning the Longformer model on Luganda and English text to detect the severity of depression using a custom-made dataset from Reddit, we have shown that the longformer model can be finetuned on both languages outperforming machine learning models, that acted as baseline models. 

We have also demonstrated that despite Luganda not being one of the languages originally trained with the Longformer model, it was successfully fine-tuned. However, we believe that training the Longformer model with the Luganda language and subsequently fine-tuning it could potentially yield better results.

The dataset used was small, and hence this is presumably the reason why the baseline machine learning models failed. A large dataset could improve the overall performance of all models and provide more samples for the severe and moderate classes.

The use of Google Translate as a machine translation model also affected the performance of the Luganda experiments. The service of a linguistic expert could be  employed for a better translation output. 

\bibliographystyle{ieeetr}
\bibliography{references}

\begin{thebibliography}{10}

\bibitem{fogarty2021physical}
F.~Fogarty, G.~McCombe, K.~Brown, T.~Van~Amelsvoort, M.~Clarke, and W.~Cullen, ``Physical health among patients with common mental health disorders in primary care in europe: a scoping review,'' {\em Irish journal of psychological medicine}, vol.~38, no.~1, pp.~76--92, 2021.

\bibitem{mcallister2020identification}
R.~McAllister-Williams, C.~Arango, P.~Blier, K.~Demyttenaere, P.~Falkai, P.~Gorwood, M.~Hopwood, A.~Javed, S.~Kasper, G.~Malhi, {\em et~al.}, ``The identification, assessment and management of difficult-to-treat depression: an international consensus statement,'' {\em Journal of Affective Disorders}, vol.~267, pp.~264--282, 2020.

\bibitem{beck1987beck}
A.~T. Beck, R.~A. Steer, G.~K. Brown, {\em et~al.}, {\em Beck depression inventory}.
\newblock Harcourt Brace Jovanovich New York:, 1987.

\bibitem{redditinc}
{Reddit Inc.}, ``{Reddit Blog: Apifacts}.'' \url{https://www.redditinc.com/blog/apifacts}.
\newblock Retrieved on August 14, 2023.

\bibitem{reddittransparency}
``{Reddit Transparency Report 2022}.'' \url{https://www.redditinc.com/policies/2022-transparency-report}.
\newblock Retrieved on August 14, 2023.

\bibitem{kim2022analysis}
N.~H. Kim, J.~M. Kim, D.~M. Park, S.~R. Ji, and J.~W. Kim, ``Analysis of depression in social media texts through the patient health questionnaire-9 and natural language processing,'' {\em Digital Health}, vol.~8, p.~20552076221114204, 2022.

\bibitem{kim2020deep}
J.~Kim, J.~Lee, E.~Park, and J.~Han, ``A deep learning model for detecting mental illness from user content on social media,'' {\em Scientific reports}, vol.~10, no.~1, p.~11846, 2020.

\bibitem{murarka2021classification}
A.~Murarka, B.~Radhakrishnan, and S.~Ravichandran, ``Classification of mental illnesses on social media using roberta,'' in {\em Proceedings of the 12th international workshop on health text mining and information analysis}, pp.~59--68, 2021.

\bibitem{naseem2022early}
U.~Naseem, A.~G. Dunn, J.~Kim, and M.~Khushi, ``Early identification of depression severity levels on reddit using ordinal classification,'' in {\em Proceedings of the ACM Web Conference 2022}, pp.~2563--2572, 2022.

\bibitem{razavi2020depression}
R.~Razavi, A.~Gharipour, and M.~Gharipour, ``Depression screening using mobile phone usage metadata: a machine learning approach,'' {\em Journal of the American Medical Informatics Association}, vol.~27, no.~4, pp.~522--530, 2020.

\bibitem{francese2023emotion}
R.~Francese and P.~Attanasio, ``Emotion detection for supporting depression screening,'' {\em Multimedia Tools and Applications}, vol.~82, no.~9, pp.~12771--12795, 2023.

\bibitem{vaswani2017attention}
A.~Vaswani, N.~Shazeer, N.~Parmar, J.~Uszkoreit, L.~Jones, A.~N. Gomez, {\L}.~Kaiser, and I.~Polosukhin, ``Attention is all you need,'' {\em Advances in neural information processing systems}, vol.~30, 2017.

\bibitem{devlin2018bert}
J.~Devlin, M.-W. Chang, K.~Lee, and K.~Toutanova, ``Bert: Pre-training of deep bidirectional transformers for language understanding,'' {\em arXiv preprint arXiv:1810.04805}, 2018.

\bibitem{shounak2022reddit}
R.~Shounak, S.~Roy, V.~Kumar, and V.~Tiwari, ``Reddit comment toxicity score prediction through bert via transformer based architecture,'' in {\em 2022 IEEE 13th Annual Information Technology, Electronics and Mobile Communication Conference (IEMCON)}, pp.~0353--0358, IEEE, 2022.

\bibitem{caselli2020hatebert}
T.~Caselli, V.~Basile, J.~Mitrovi{\'c}, and M.~Granitzer, ``Hatebert: Retraining bert for abusive language detection in english,'' {\em arXiv preprint arXiv:2010.12472}, 2020.

\bibitem{gabin2021multiple}
J.~Gab{\'\i}n, A.~P{\'e}rez, and J.~Parapar, ``Multiple-choice question answering models for automatic depression severity estimation,'' {\em Engineering Proceedings}, vol.~7, no.~1, p.~23, 2021.

\bibitem{liu2019roberta}
Y.~Liu, M.~Ott, N.~Goyal, J.~Du, M.~Joshi, D.~Chen, O.~Levy, M.~Lewis, L.~Zettlemoyer, and V.~Stoyanov, ``Roberta: A robustly optimized bert pretraining approach,'' {\em arXiv preprint arXiv:1907.11692}, 2019.

\bibitem{beltagy2020longformer}
I.~Beltagy, M.~E. Peters, and A.~Cohan, ``Longformer: The long-document transformer,'' {\em arXiv preprint arXiv:2004.05150}, 2020.

\bibitem{li2023comparative}
Y.~Li, R.~M. Wehbe, F.~S. Ahmad, H.~Wang, and Y.~Luo, ``A comparative study of pretrained language models for long clinical text,'' {\em Journal of the American Medical Informatics Association}, vol.~30, no.~2, pp.~340--347, 2023.

\bibitem{owen2023enabling}
D.~Owen, D.~Antypas, A.~Hassoulas, A.~F. Pardi{\~n}as, L.~Espinosa-Anke, J.~C. Collados, {\em et~al.}, ``Enabling early health care intervention by detecting depression in users of web-based forums using language models: Longitudinal analysis and evaluation,'' {\em JMIR AI}, vol.~2, no.~1, p.~e41205, 2023.

\bibitem{sheu2023ai}
Y.-h. Sheu, C.~Magdamo, M.~Miller, S.~Das, D.~Blacker, and J.~W. Smoller, ``Ai-assisted prediction of differential response to antidepressant classes using electronic health records,'' {\em NPJ Digital Medicine}, vol.~6, no.~1, p.~73, 2023.

\bibitem{iversen2021child}
S.~A. Iversen, J.~Nalugya, J.~N. Babirye, I.~M.~S. Engebretsen, and N.~Skokauskas, ``Child and adolescent mental health services in uganda,'' {\em International journal of mental health systems}, vol.~15, pp.~1--12, 2021.

\bibitem{newman2022access}
M.~W. Newman, M.~Hawrilenko, M.~Jakupcak, S.~Chen, and J.~C. Fortney, ``Access and attitudinal barriers to engagement in integrated primary care mental health treatment for rural populations,'' {\em The Journal of Rural Health}, vol.~38, no.~4, pp.~721--727, 2022.

\bibitem{dagsvold2015can}
I.~Dagsvold, S.~M{\o}llersen, and V.~Stordahl, ``What can we talk about, in which language, in what way and with whom? sami patients' experiences of language choice and cultural norms in mental health treatment,'' {\em International journal of circumpolar health}, vol.~74, no.~1, p.~26952, 2015.

\bibitem{subredditstats}
``r/depression statistics.'' \url{https://subredditstats.com/r/depression}.
\newblock Retrieved on August 15, 2023.

\bibitem{loper2002nltk}
E.~Loper and S.~Bird, ``Nltk: The natural language toolkit,'' {\em arXiv preprint cs/0205028}, 2002.

\bibitem{lewis2019bart}
M.~Lewis, Y.~Liu, N.~Goyal, M.~Ghazvininejad, A.~Mohamed, O.~Levy, V.~Stoyanov, and L.~Zettlemoyer, ``Bart: Denoising sequence-to-sequence pre-training for natural language generation, translation, and comprehension,'' {\em arXiv preprint arXiv:1910.13461}, 2019.

\bibitem{chawla2002smote}
N.~V. Chawla, K.~W. Bowyer, L.~O. Hall, and W.~P. Kegelmeyer, ``Smote: synthetic minority over-sampling technique,'' {\em Journal of artificial intelligence research}, vol.~16, pp.~321--357, 2002.

\bibitem{basafa2021nlp}
H.~Basafa, S.~Movahedi, A.~Ebrahimi, A.~Shakery, and H.~Faili, ``Nlp-iis@ ut at semeval-2021 task 4: Machine reading comprehension using the long document transformer,'' {\em arXiv preprint arXiv:2105.03775}, 2021.

\bibitem{jain2022hugging}
S.~M. Jain, ``Hugging face,'' in {\em Introduction to Transformers for NLP: With the Hugging Face Library and Models to Solve Problems}, pp.~51--67, Springer, 2022.

\end{thebibliography}
\vspace{12pt}

\end{document}